\documentclass[10pt,twocolumn,letterpaper]{article}

\usepackage[pagenumbers]{cvpr} % To force page numbers, e.g. for an arXiv version

\usepackage[accsupp]{axessibility}

\usepackage{graphicx}
\usepackage{amsmath}
\usepackage{amssymb}
\usepackage{booktabs}
\usepackage{dblfloatfix}

\usepackage[pagebackref,breaklinks,colorlinks]{hyperref}

\usepackage[capitalize]{cleveref}
\crefname{section}{Sec.}{Secs.}
\Crefname{section}{Section}{Sections}
\Crefname{table}{Table}{Tables}
\crefname{table}{Tab.}{Tabs.}

\def\cvprPaperID{50} % *** Enter the EarthVision Paper ID here (taken from CMT)
\def\confName{EarthVision}
\def\confYear{2023}

\begin{document}

% \title{Inferring the past: a CNN--LSTM satellite fusion framework for historical inundation}
% to fuse sattelites for historical inundation mapping
\title{Inferring the past: a combined CNN--LSTM deep learning framework to fuse satellites for historical inundation mapping}

\author{Jonathan Giezendanner\\
University of Arizona\\
87521 Tucson, USA\\
{\tt\small \url{https://jgiezendanner.com}}
\and
Rohit Mukherjee\\
University of Arizona\\
87521 Tucson, USA\\
{\tt\small \url{http://rohitmukherjee.space}}
\and
Matthew Purri\\
University of Arizona\\
87521 Tucson, USA\\
{\tt\small \url{mpurri@arizona.edu}}
\and
% Jonathan Sullivan\\
% University of Arizona\\
% 87521 Tucson, USA\\
% {\tt\small \url{jasullivan@arizona.edu}}
% \and
Mitchell Thomas\\
Columbia University\\
10964 Palisades, USA\\
{\tt\small \url{mlt2177@columbia.edu}}
\and
Max Mauerman\\
Columbia University\\
10964 Palisades, USA\\
{\tt\small \url{mm5330@columbia.edu}}
\and
% Upmanu Lall\\
% Columbia University\\
% 10964 Palisades, USA\\
% {\tt\small \url{ula2@columbia.edu}}
% \and
A.K.M. Saiful Islam\\
Bangladesh University of Engineering and Technology\\
1000 Dhaka, Bangladesh\\
{\tt\small \url{akmsaifulislam@iwfm.buet.ac.bd}}
\and
Beth Tellman\\
University of Arizona\\
87521 Tucson, USA\\
{\tt\small \url{btellman@arizona.edu}}
}
\maketitle

\begin{abstract}
    Mapping floods using satellite data is crucial for managing and mitigating flood risks.
Satellite imagery enables rapid and accurate analysis of large areas, providing critical information for emergency response and disaster management.
Historical flood data derived from satellite imagery can inform long-term planning, risk management strategies, and insurance-related decisions.
The Sentinel-1 satellite is effective for flood detection, but for longer time series, other satellites such as MODIS can be used in combination with deep learning models to accurately identify and map past flood events.
We here develop a combined CNN--LSTM deep learning framework to fuse Sentinel-1 derived fractional flooded area with MODIS data in order to infer historical floods over Bangladesh.
The results show how our framework outperforms a CNN-only approach and takes advantage of not only space, but also time in order to predict the fractional inundated area.
The model is applied to historical MODIS data to infer the past 20 years of inundation extents over Bangladesh and compared to a thresholding algorithm and a physical model.
Our fusion model outperforms both models in consistency and capacity to predict peak inundation extents.
    \vspace{1cm}
\end{abstract}

\section{Introduction}
\label{sec:intro}
Mapping floods supports building resilience and the adoption of mitigation strategies to minimize losses in current and future events.
Timely and accurately identifying regions affected by flooding is crucial for emergency responses and disaster management efforts.
Satellite imagery provides a unique vantage point for near real time mapping, especially for remote and inaccessible regions.

Satellite data is an effective tool for mapping floods as it allows for rapid and accurate analysis of large areas, even in remote or inaccessible regions.
With satellite imagery, flood-affected areas can be monitored and assessed in near-real-time, providing critical information for emergency response and disaster management efforts.

Additionally, satellite data can provide historical information that can be used to analyze trends and patterns in flooding over time, helping to inform long-term resiliency planning and risk management strategies.
Historical flood data derived from satellite imagery can be an essential resource for understanding flood risk over time, particularly for insurance-related cases~\cite{tellman2022a}.
By analysing historical data, insurers can better assess the likelihood and potential severity of future flood events, which can inform risk management decisions, pricing, and coverage options.

Traditional flood insurance for financial protection from flood damages involves contracts where adjusters visit individual homes to assess damages and determine corresponding payouts.
However, traditional insurance is often unavailable in regions of the world most vulnerable to floods, such as Bangladesh.
Flood index insurance is an affordable and scaleable alternative whereby a data source, often from weather stations or satellites, generates a “flood index” (\eg, inundated area) that should correlate with the damage insured \cite{surminski2016}.
This makes insurance scalable in remote areas where traditional insurance penetration is low.
Public satellite sensors such as MODIS are currently being piloted for flood index insurance in Southeast Asia~\cite{matheswaran2019} and Colombia~\cite{gallinluke2022}.
MODIS is used because its 20+ year history makes it possible to estimate probability of exceedance needed to both price an insurance instrument and determine an appropriate trigger for payout~\cite{tellman2022a}.
Additionally, MODIS is available on a daily basis, and its derived 8-days composite image provides a relatively cloud free image.
Other products like Landsat, although available for a longer time, do not provide such a consistent image series (low revisit frequency, smaller extent, harmonisation issues between missions and instrument failure amongst other challenges).

MODIS is an optical sensor with two images acquired daily, and is commonly used for global and near real time flood mapping ~\cite{tellman2021}.
%TODO: cite more
However, clouds often obscure the view in optical sensors, making it difficult to map floods in many cases.
Furthermore, the coarse spatial resolution of 250m, 500m, and 1km of MODIS makes it unsuitable for accurately extracting flooded features in complex and heterogeneous landscapes such as urban regions.
Together, the coarse spatial resolution and cloud cover can lead to basis risk, where the data for index insurance is not correlated with damage on the ground.
Recent evaluations of using MODIS for inundation maps in Bangladesh show low to no correlation with reported damage, raising concerns that insurance product based on this sensor could cause basis risk~\cite{thomas2023}.
To overcome the limitations of optical sensors, radar sensors~\cite{kang2018, hostache2018, herrera-cruz} detecting water signals through clouds consistently  are preferable for insurance applications~\cite{uddin2019, colosio2022}.

Sentinel-1 is an ideal satellite for flood detection due to its radar sensor that provides high-resolution (10 meters) images of the Earth's surface, even in cloudy conditions, making it suitable for accurate and consistent flood extent mapping.
However, it is only reliably available since 2017 and is therefore unsuitable for historical flood mapping.
For applications where longer time series are needed, such as return period estimates that require at least 20 years of data \cite{benami2021}, other satellites have to be considered.

Fusing together Sentinel-1 with a longer MODIS time series is a potential solution to overcome the disadvantages of either sensor for index-based insurance and leverage the advantages of high resolution modern flood detection and historical data availability.

We propose a deep learning model trained on Sentinel-1 derived fractional inundated area to predict historical MODIS satellite data.
We utilize a Convolutional Neural -- Long Short-Term Memory Network (CNN--LSTM) to take advantage of both the spatial and temporal feature representation of inundation and compare it to a traditional Convolutional Neural Network (CNN).
We apply the CNN--LSTM model from 2000~-~2021 over all of Bangladesh at 500m resolution, to demonstrate the potential of our fusion model to be used in index insurance pricing applications.

\section{Methods}
\label{sec:methods}
\begin{figure}[tbh]
    \centering
    \includegraphics[width=.8\linewidth]{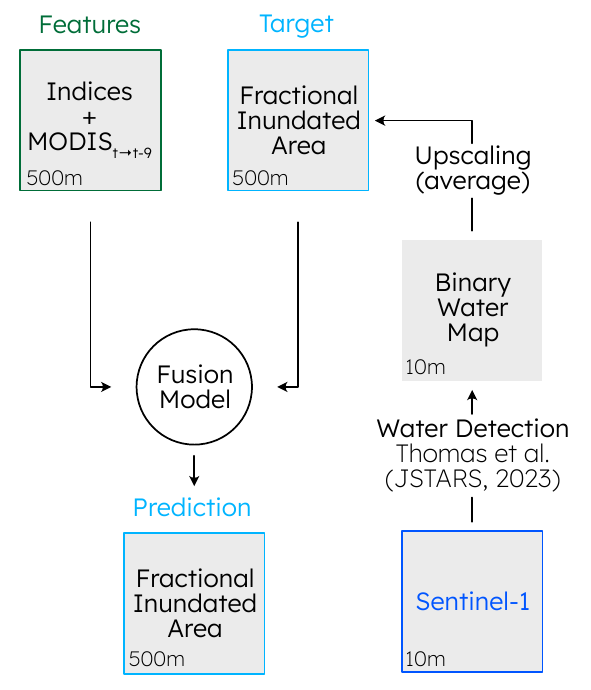}

    \caption{
        Modelling framework:
        The target flood maps are generated from Sentinel-1 data.
        Water is detected with a dynamic thresholding algorithm tailored to Bangladesh and then upscaled (averaged) to MODIS resolution.
        The fusion model uses hydrologically relevant indices (elevation, slope, and height above nearest drainage (HAND)) and a time series of 10 MODIS Terra 8-day composite images to regress the target fractional inundated area.}
    \label{fig:methods:framework}
\end{figure}

We train a deep learning fusion model (\cref{fig:methods:framework}) to regress fractional inundated area based on an 8-day MODIS composite of satellite imagery and static landscape features (elevation, slope and height above nearest drainage (HAND)).
The inundated labels are derived from 10 meters resolution Sentinel-1 imagery.
The fractional inundated area is the percentage of area covered with water.

Our proposed deep learning framework combines the spatial encoding ability of CNNs with the temporal encoding of LSTMs to form a CNN--LSTM (not to be confused with Convolutional LSTMs~\cite{boulila2021}).
CNNs are commonly used in computer vision applications~\cite{lecun2015, wu2019, ajit2020}, and numerous applications show how they can be used in the context of satellite based flood detection~\cite{nearing2021, shen2018, bai2021, bonafilia2020}.
LSTMs on the other hand have proven particularly powerful in sequential time series applications \cite{ahmed2022}.
In hydrology in particular, LSTMs have been applied in the context of rainfall-runoff predictions~\cite{frame2022,kratzert2019, frame2021, frame2020, nearing2020,kratzert2018} and flood forecasting~\cite{lexuan-hien2019, ding2020, song2019}.
Our motivation for using LSTMs over other ways of capturing time (\eg, transformers, stacked image time series as CNN features), is physically informed.
LSTMs conserve the sequence of the time series, whereas transformers reshuffle the pixel stack~\cite{katrompas2022}, and CNNs arbitrarily link the features together.
LSTMs should thus capture the physical dynamic of the inundation in time. 
Additionally, transformers are very data hungry~\cite{xu2021}, especially when training them on a problem that is very different from the problem they were designed for~\cite{ranftl2021}, which makes their use unreasonable for this proof of concept paper.
Future work should explore these alternatives.
CNN-LSTMs have been applied in various remote sensing contexts~\cite{interdonato2019, sun2019a, wang2021a}, but, although applications of CNN--LSTMs exist in hydrology~\cite{wu2020, li2022b}, to the best of our knowledge, CNN--LSTMs have not been used to regress satellite based inundation estimates, where the time series of satellite imagery is the main driver of the model.

\subsection{Data}
    \begin{figure}[tbh]
    \centering
    \includegraphics[width=1\linewidth]{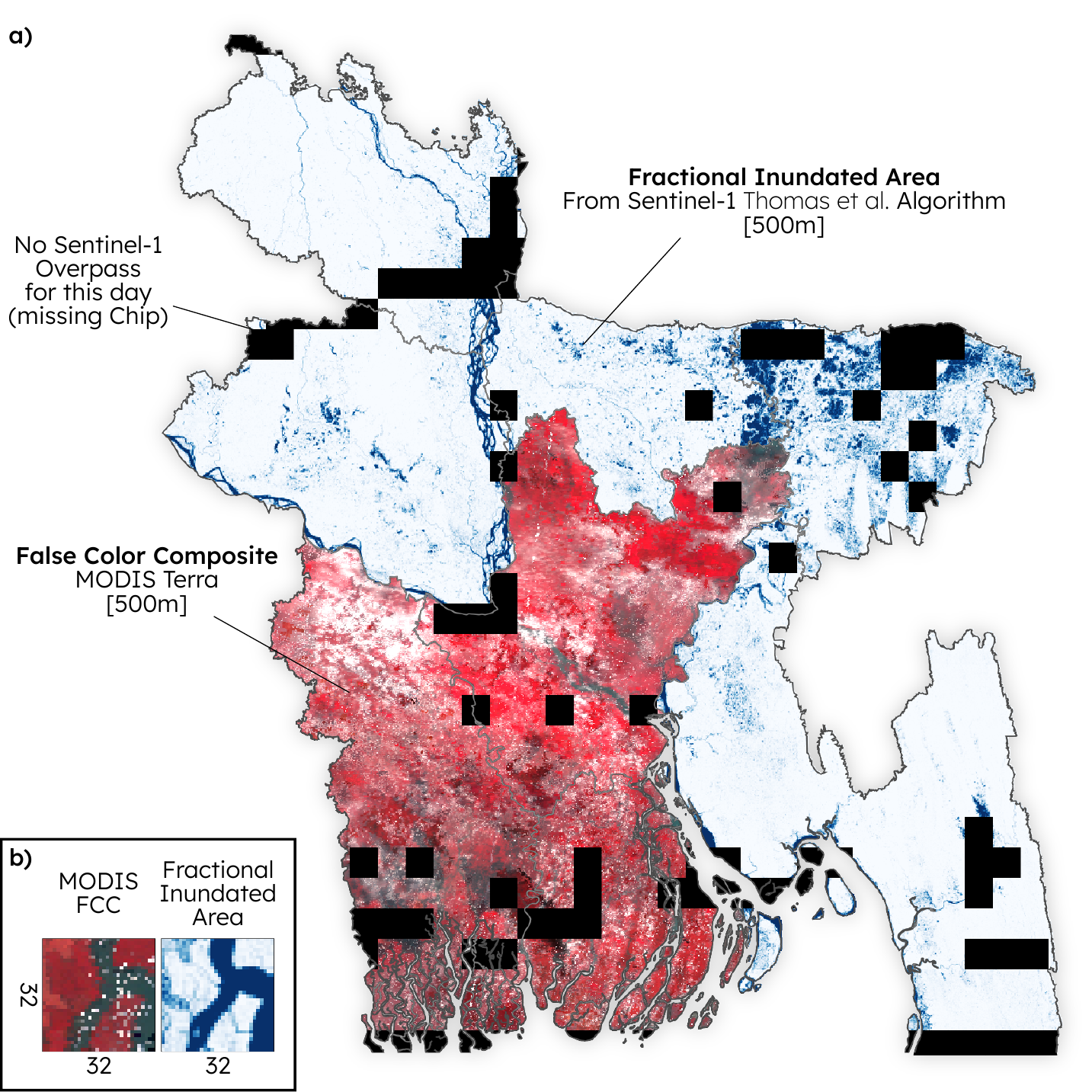}

    \caption{Example of data collected and used for this modelling framework:
    a) example of target fractional inundated area for a chosen week (week of August 9, 2021) upscaled from binary Sentinel-1 water maps, along a False Color Composite (FCC: near-infrared (NIR), red, blue) image of MODIS for the same date.
    The black chips indicate a chip where at least one pixel is missing for this week at this place (if more than one pixel is missing, the chip is discarded, due to missing sentinel overpass during the 8 days).
    Note that the shown data overlap between Sentinel-1 and MODIS is specific for the chosen week, but varies in time.
    b) zoom in on a single 32~x~32 chip with MODIS FCC on the left, and Fractional Inundated Area on the right.
    The resolution for both a) and b) is 500 meters.
    }
    \label{fig:methods:dataOverview}
\end{figure} 
    \subsubsection{Target Fractional Inundated Area Maps}
        We generate fractional flooded area at a 500 meters resolution.
        The target resolution is chosen to match MODIS Terra's 500 meter resolution.
        We first generate binary flood maps at the Sentinel-1 native resolution (10 m).
        We apply a dynamic thresholding method derived from DeVries \etal~\cite{devries2020} specifically developed and verified over Bangladesh~\cite{thomas2023}.
        
        This Sentinel-1 algorithm identifies a dry-season baseline from a set of VV and VH backscatter images taken during periods of low soil moisture.
With these dry season baseline images, the pixel-wise mean and standard deviation images are calculated.
From these, a z-score image (number of standard deviations from the mean, unitless) is computed for each Sentinel-1 image in the historical record.
Surface water is defined as any pixel with a Z-score below -2 in either VV or VH bands, or by VH backscatter below -27 dB.
The map is then smoothed by a majority filter at 30 meters to remove SAR noise.
The final result of this algorithm is a binary surface water map at 10 meters resolution available every $\sim$5 days on average.

        Based on this binary inundation map, we calculate the fractional inundated area by upsampling the 10 meter resolution map to 500 meters. 
        Each 500 m x 500 m grid cell thus represents the average number of pixels where water is detected over 2,500 pixels.
        We extract images at a chip size of 32x32 to allow the network to learn spatial relationships, and permit a reasonable overlap between the Sentinel-1 extent and MODIS resolution.
        Larger chips would benefit the CNN and allow for a better spatial context, but matching large MODIS and Sentinel-1 extents appears very difficult.
        Each 32x32 image patch contains information from 1,600~x~1,600 Sentinel-1 pixels or covers an area of 256 km$^2$.

        We extract all 32~x~32, 500 meters resolution, chips for the years 2017 to 2021 (5 years) in a regular grid over Bangladesh (\cf \cref{fig:methods:dataOverview}). With this method, we generate a total of 372,263 target chips covering all of Bangladesh, over 5 years, with a 2~-~10 days interval between images (Sentinel-1 revisitation time).
        Out of those, 150,946 had no missing Sentinel-1 data (complete chip) and were used in the context of this study.

    \subsubsection{Features}
        We use a combination of dynamic (satellite remote sensing data) and static (derived from digital elevation models) features to regress the target fractional inundated area, for a total of 10 features.
        Each feature stack is stored in a 32~x~32, 500 meters resolution raster chip.

        \textit{Dynamic Features:} For each chip, we select the 8-Day Global 500 meters MODIS Terra Surface Reflectance image (MOD09A1.061) that overlaps the Sentinel-1 collection date.
        We use the 7 first bands (red, near-infrared (NIR), blue, green, SWIR 1-3), which are min-max normalised between 0 and 1 based on the sensor's data range (-100, 16,000).
        We also extract the 9 previous MODIS images, to be used in the LSTM (see below), for a total of 10 images for each Sentinel-1 date.
        This period of approximately 2.5 months covers a majority of the inundation peak in Bangladesh.
        Future work should consider extending this time series to a full year of data to capture yearly dynamics, although this would come to the cost of high computational resources.

        \textit{Static Features:} Hydrologically relevant features and indexes have been shown to improve model performance in flood related applications~\cite{schumann2021}.
        We here complement the MODIS data with elevation, slope and height above nearest drainage (HAND~\cite{nobre2011}).
        Elevation is given as the re-sampled (downsampled/upscsaled using the mean from 30 meters to 500 meters) elevation in each 500m pixel derived from the Digital Elevation Model (DEM) product FABDEM~\cite{hawker2022}.
        The output value is min-max normalised between 0 and 1 with a reasonable elevation range for Bangladesh (0, 100).
        Slope is first calculate based on the FABDEM product, and then re-sampled (mean) and expressed as its hyper tangent to bound the values between 0 and 1. 
        HAND is extracted from Yamazaki \etal~\cite{yamazaki2019}, the logarithm is applied to reduce the data spread, and after re-sampling (mean) a min-max normalisation is applied to bound the values between 0 and 1.

\subsection{Model}
    As mentioned above, we propose a deep learning framework that combines CNNs and LSTMs sequentially (\cf \cref{fig:methods:architecture}).
    For the particular implementation of CNN--LSTM we propose, for each time step, the features are initially passed through the first CNN (CNN A in \cref{fig:methods:architecture}) to summarize the multi-band images into a single band.
    This output generates, for each pixel in the 32~x~32 chip, a time series of spatially contextualised data.
    Note that, although each group of features (time step) is passed through CNN A independently, CNN A is the same network across groups and the weights are shared.
    CNN A is comprised of 4 convolution layers (Conv2D $\rightarrow$ BatchNorm2D $\rightarrow$ ReLU) that inflates the number of layers from 10 to 128, and one final Conv2D layer which brings it down to 1 layer just before the LSTM input.
    The LSTM is then fed, for each pixel, a time series consisting of the 9 encoded values.

    \begin{figure}[tbh]
    \centering
    \includegraphics[width=1\linewidth]{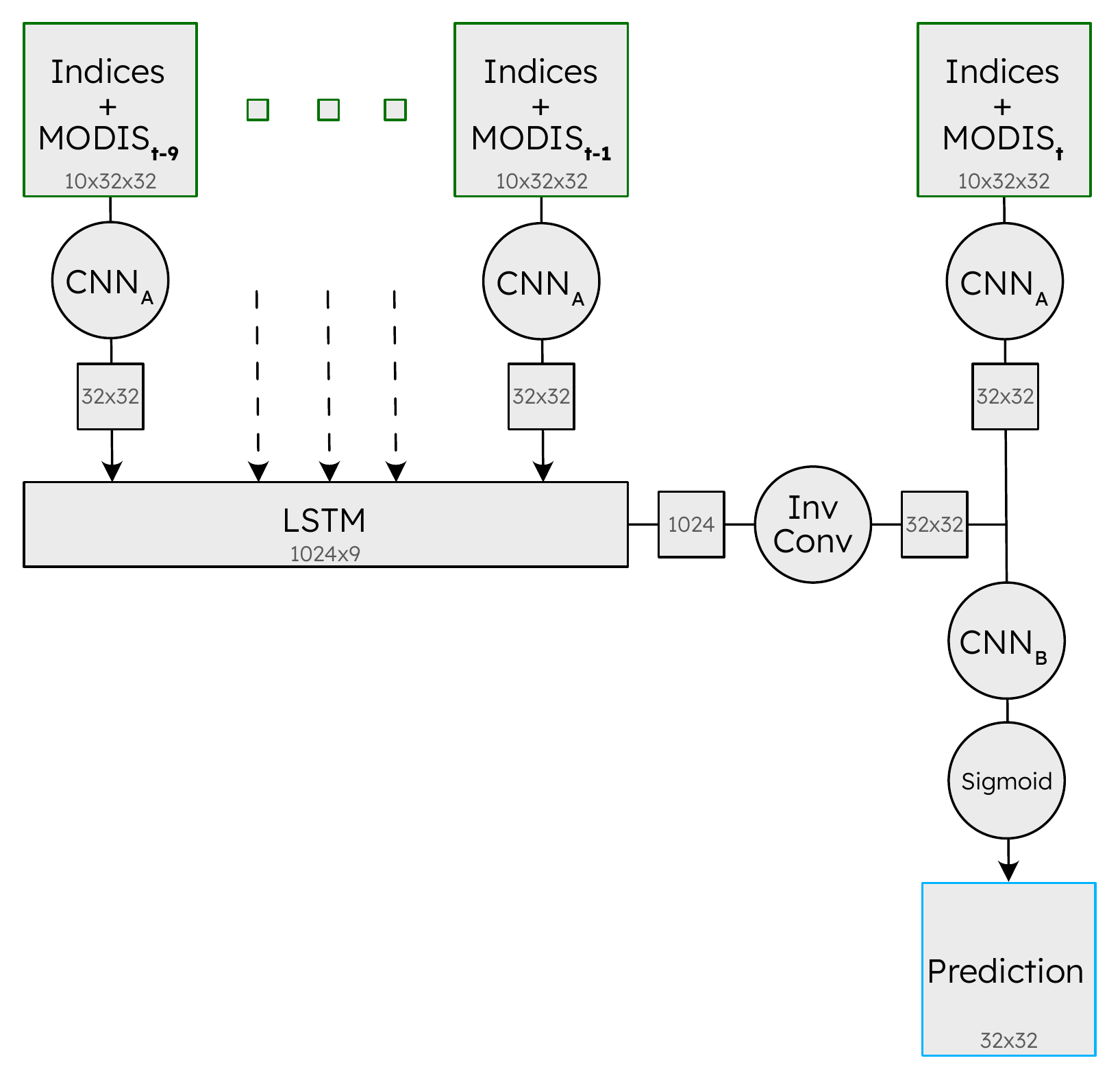}

    \caption{
        Schematic representation of the deep learning architecture: 
        A CNN (A) is run on each time step of a 10 input MODIS images and hydrologically relevant indices, for a total of 10 input bands per time step.
        The CNN outputs of the 9 time steps before time $t$ is passed through a LSTM network, where each of the 1'024 pixels is considered as an independent time series.
        The output of the LSTM is passed through an inverse convolution layer, and ,concatenated with the output of CNN A at time step $t$, passed through a second CNN (B).
        Finally, a sigmoid is applied to bound the prediction between 0 and 1.
        Note that CNN A is the same network for all time steps and that the weights are shared, whereas CNN B is different.
        The CNN's architecture is described in further detail in the text.
        Also note that the light gray numbers in the figure denote the different layer's dimensions.
    }
    \label{fig:methods:architecture}
\end{figure}

    The output of the LSTM is converted from a one-dimensional (1~x~1024) layer to a square image (32~x~32) with a single inverse convolutional layer, and then merged with the CNN output of the target time, to then pass through one last CNN (CNN B in \cref{fig:methods:architecture}).
    CNN B is composed of a single Conv2D layer.
    Finally, the output is passed through a sigmoid activation layer to bound the prediction between 0 and 1.

    Both CNNs use a convolution scheme with a kernel size of 3, reflected padding, and a stride of 1.
    The input image size is retained throughout the convolution layers (32~x~32).
    This choice was motivated by the size of the image.

\subsubsection{Training and Cross-Validation}\label{sec:methods:model:training}
    The model is trained with a leave-one-out cross-validation scheme, where each of the 5 years is withheld for validation one at a time, \ie, the model is trained 5 times with varying number of chips (\cf \cref{tab:methods:numberchips}).
    \begin{table}[h]
    \centering
    \begin{tabular}{@{}cc@{}}
        \toprule
        Leave-out Year & Number of chips \\
        \midrule
        2017 & 24,845\\
        2018 & 21,558\\
        2019 & 17,279\\
        2020 & 43,815\\
        2021 & 43,449\\
        \midrule
        Total & 150,946\\
        \bottomrule
    \end{tabular}
    \caption{
        Number of complete chips collected over Bangladesh for each year.
        Only complete chips are considered here, chips where parts of the Sentinel-1 data was missing are not considered.
    }
    \label{tab:methods:numberchips}
\end{table}

% \begin{table}[h]
%     \centering
%     \begin{tabular}{@{}l|ccccc@{}}
%         \toprule
%         Year &  2017 & 2018 & 2019 & 2020 & 2021\\
%         % \midrule
%         Chips & 24'845 & 21'558 & 17'279 & 43'815 & 43'449\\
%         \bottomrule
%     \end{tabular}
%     \caption{Number of chips collected over Bangladesh for each year.}
%     \label{tab:methods:numberchips}
% \end{table}

    For each cross-validated year, we train the model on 20 epochs with a learning rate of 10$^{\text{-3}}$, then 5 epochs with a learning rate of 10$^{\text{-4}}$ and finally 5 epochs with a learning rate of 10$^{\text{-5}}$.
    We use root mean square error (RMSE) for the loss function and the Ranger optimiser.
    All models reached convergence of loss after the 30 epochs.
    We use rotation, flip and dihedral image transformation to augment the data.

\subsubsection{Model Baseline}
    In order to assess the performance of the proposed model, we compare it to a CNN-only model without temporal data.
    We train a CNN with the same architecture as CNN A (\cf \cref{fig:methods:architecture}) with the same cross-validation and training scheme described in \cref{sec:methods:model:training} and compare it to the combined CNN--LSTM network.

\section{Results}
\label{sec:res}
\subsection{Capturing the Temporal Evolution in the Deep Learning Model}
    It is important to assess whether including temporal information improves the prediction of fractional inundated area or if the spatial context is sufficient.
    The CNN model is trained with the same set of hyperparameters and cross-validation scheme as the CNN--LSTM, and the R$^{\text{2}}$ values are reported in \cref{tab:res:stats}.
    \begin{table}[h]
    \centering
    \begin{tabular}{@{}ccc@{}}
        \toprule
        Year &  CNN & CNN--LSTM\\
        \midrule
        2017 & 0.62 &  0.66\\
        2018 & 0.55 &  0.62\\
        2019 & 0.55 &  0.67\\
        2020 & 0.63 &  0.72\\
        2021 & 0.60 &  0.67\\
        \bottomrule
    \end{tabular}
    \caption{Comparison between R$^{\text{2}}$ values calculated on leave-out year for a pure CNN model and the deep learning CNN--LSTM model presented here.
    The R$^{\text{2}}$ values are calculated over all Bangladesh.
    The CNN--LSTM model consistently outperforms the pure CNN model on all years.
    }
    \label{tab:res:stats}
\end{table}

% \begin{table}[h]
%     \centering
%     \begin{tabular}{@{}l|ccccc@{}}
%         \toprule
%         Year &  2017 & 2018 & 2019 & 2020 & 2021\\
%         % \midrule
%         R$^{\text{2}}$ (CNN) & 0.62 & 0.55 & 0.55 & 0.63 & \\
%         R$^{\text{2}}$ (CNN--LSTM) & 0.66 & 0.62 & 0.67 & 0.72 & 0.67\\
%         \bottomrule
%     \end{tabular}
%     \caption{Comp.}
%     \label{tab:res:stats}
% \end{table}

    The CNN--LSTM consistently outperforms the pure CNN network on all leave-out years.
    \cref{fig:res:CNNComp} compares, for a selected chip, the target, baseline and proposed model's temporal evolution of the fractional inundated area, and, at peak monsoon season, the inundation extent.
    \begin{figure}[htb]
    \setcounter{figure}{3}
    \centering
    \includegraphics[width=1\linewidth]{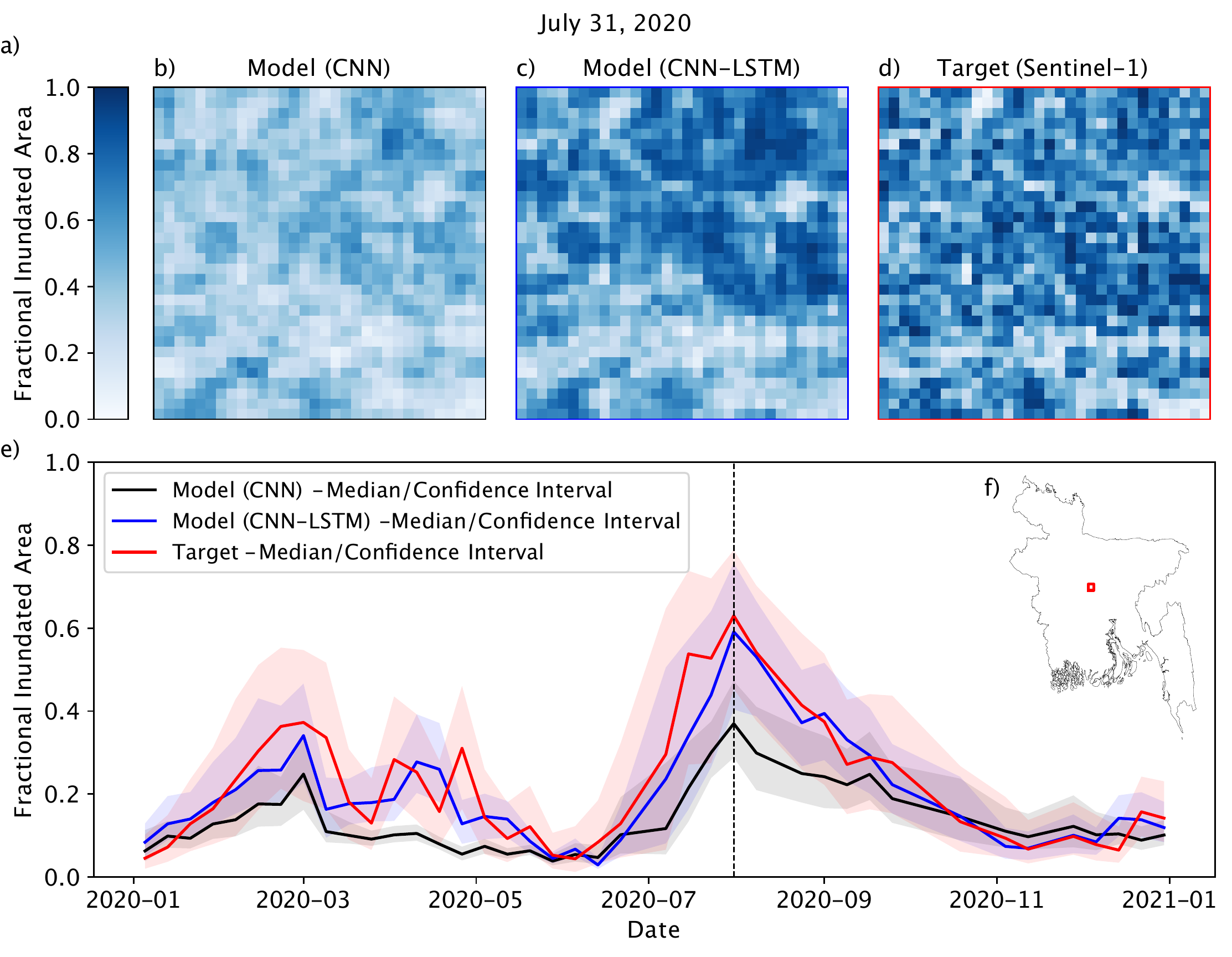}

    \caption{
        Example for a single chip (red square on the map insert (f)) of the different performance of validation data of pure CNN (chip b and black line in e) compared to the CNN--LSTM model(chip c and blue line in e), with the Sentinel-1 derived target (chip d and red line in e) as comparison.
        The model with the 2020 leave-out year is used in this example.
        The three chips (a) show the fractional inundated area at the flood peak for the monsoon season in year 2020 (at the black dashed line in the time series (e)).
        The CNN--LSTM model outperforms the pure CNN model both in temporal and spatial patterns. 
    }
    \label{fig:res:CNNComp}
\end{figure}
    The spatial patterns and inundation intensity at peak are closer matched by the CNN--LSTM than the pure CNN model.
    The temporal trend as well is closely matched by the CNN--LSTM, whereas the pure CNN misses peaks in spring, and grossly underestimates the peak inundation during the monsoon.
\begin{figure*}[bp!]
    \setcounter{figure}{5}
    \centering
    \includegraphics[width=1\linewidth]{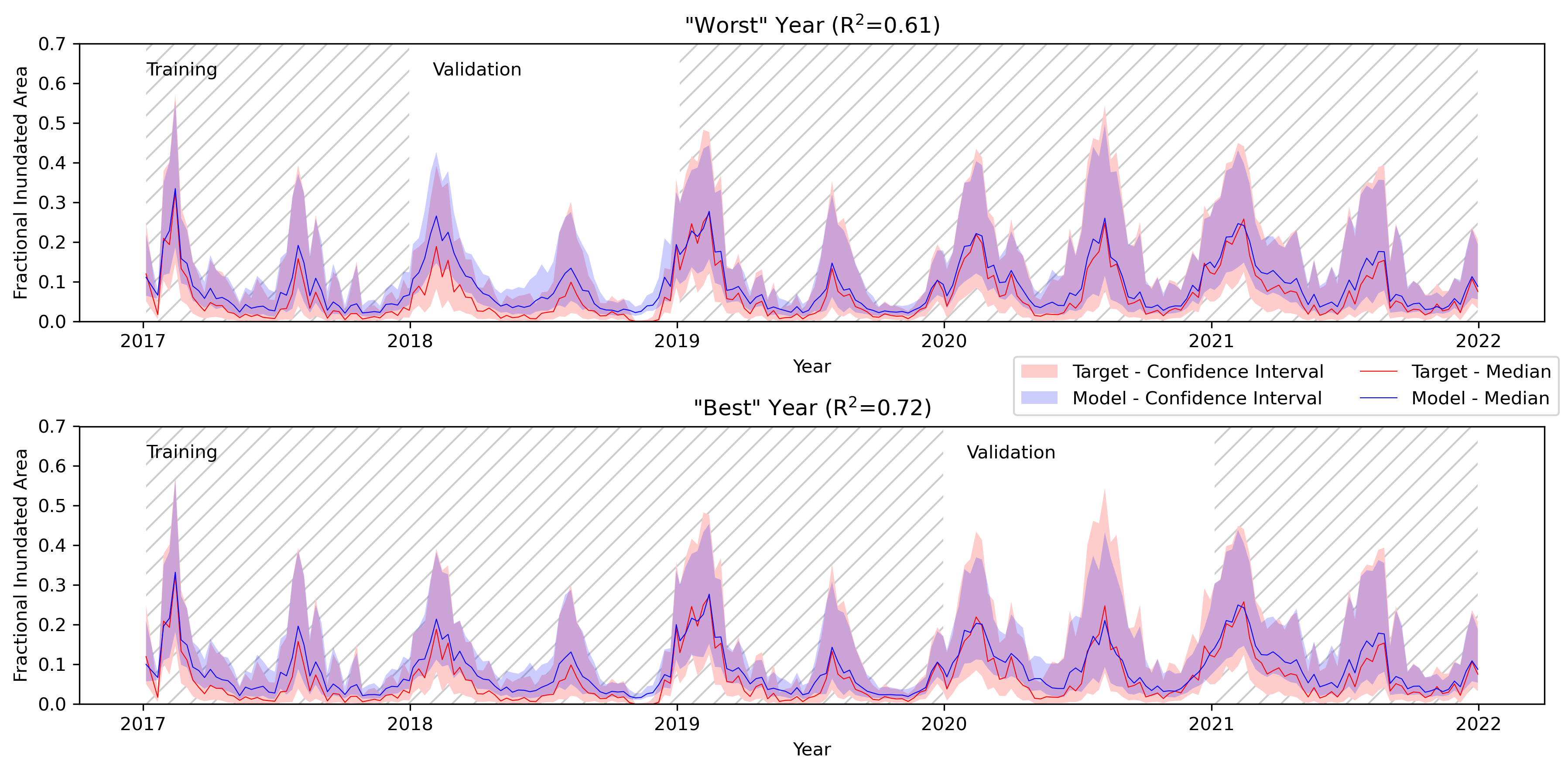}

    \caption{
        Time series of median (with confidence interval) fractional inundated area over Bangladesh for the ``worst'' (2018) and ``best'' (2020) cross-validation models (``best'' and ``worst'' in terms of R$^\text{2}$).
        The hatched area represents the data used for training, and the blank are for the validation on the cross-validation scheme.
    }
    \label{fig:res:BangladeshWorstBestYearTimeSeries}
\end{figure*}
\begin{figure}[tp]
    \setcounter{figure}{4}
    \centering
    \includegraphics[width=1\linewidth]{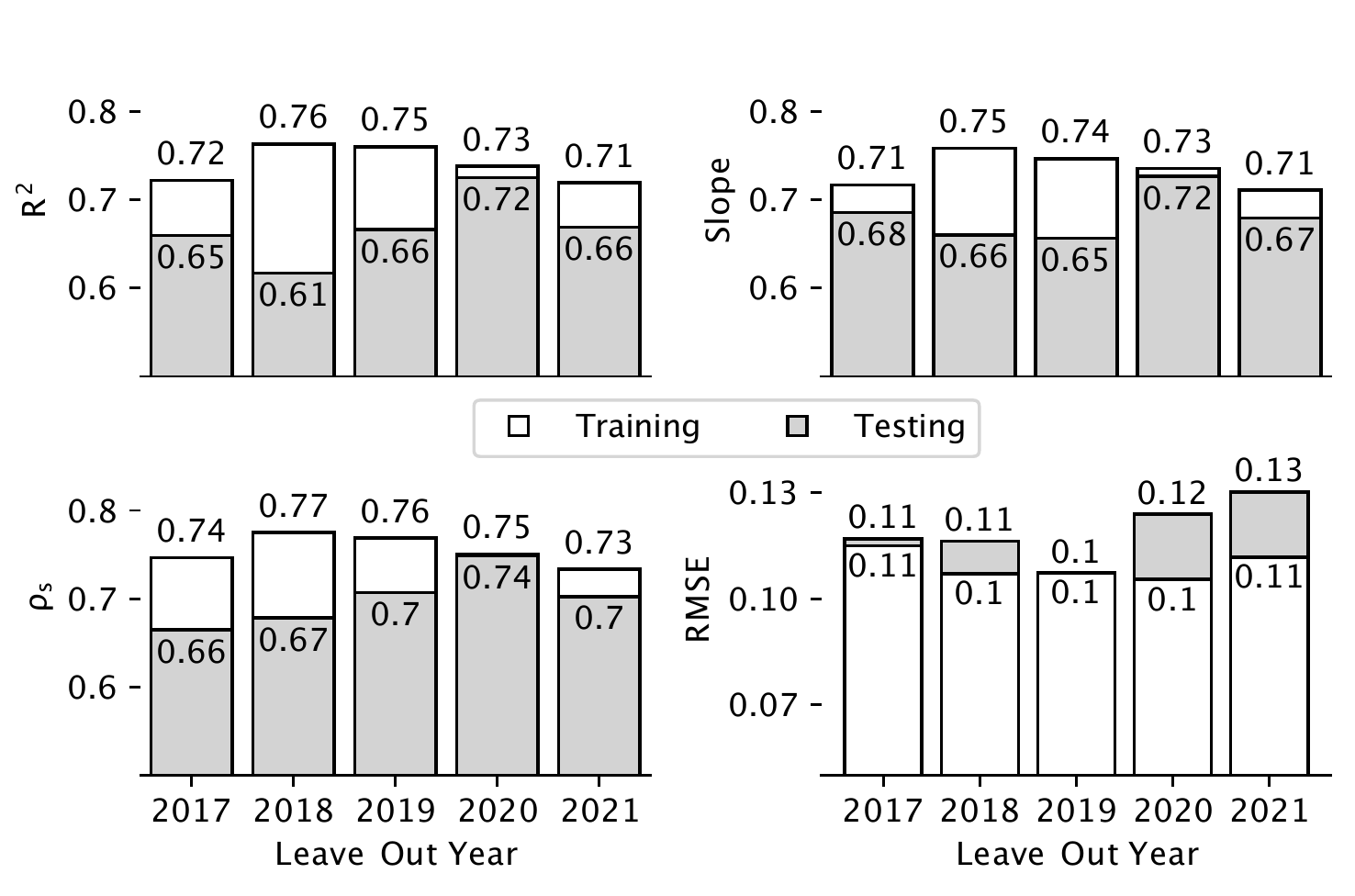}

    \caption{
        Statistical summary of the cross-validation scheme for training (white) and testing(grey).
        The years represent the year left out in the cross-validation scheme (\ie all other years are used for training).
        The figure shows four statistics: R$^\text{2}$, slope, Spearman rank-order correlation coefficient ($\rho_s$) and the root mean square error (RMSE).
        The training id displayed along side the testing to root out over-fitting.
        The average testing scores over all years are 0.66 (R$^\text{2}$), 0.68 (slope), 0.69 ($\rho_s$) and 0.1 (RMSE).
    }
    \label{fig:res:stats}
\end{figure}

% \begin{figure*}[htb]
%     \centering
%     \includegraphics[width=1\linewidth]{Figures/Results/StatsHorizontal.pdf}

%     \caption{}
%     \label{fig:res:stats}
% \end{figure*}
    Overall, the CNN-LSTM model seems to capture the temporal trends, whereas the pure CNN misses the peaks and fails to reproduce the lows during the dry season.

\begin{figure}[tbh]
    \setcounter{figure}{6}
    \centering
    \includegraphics[width=1\linewidth]{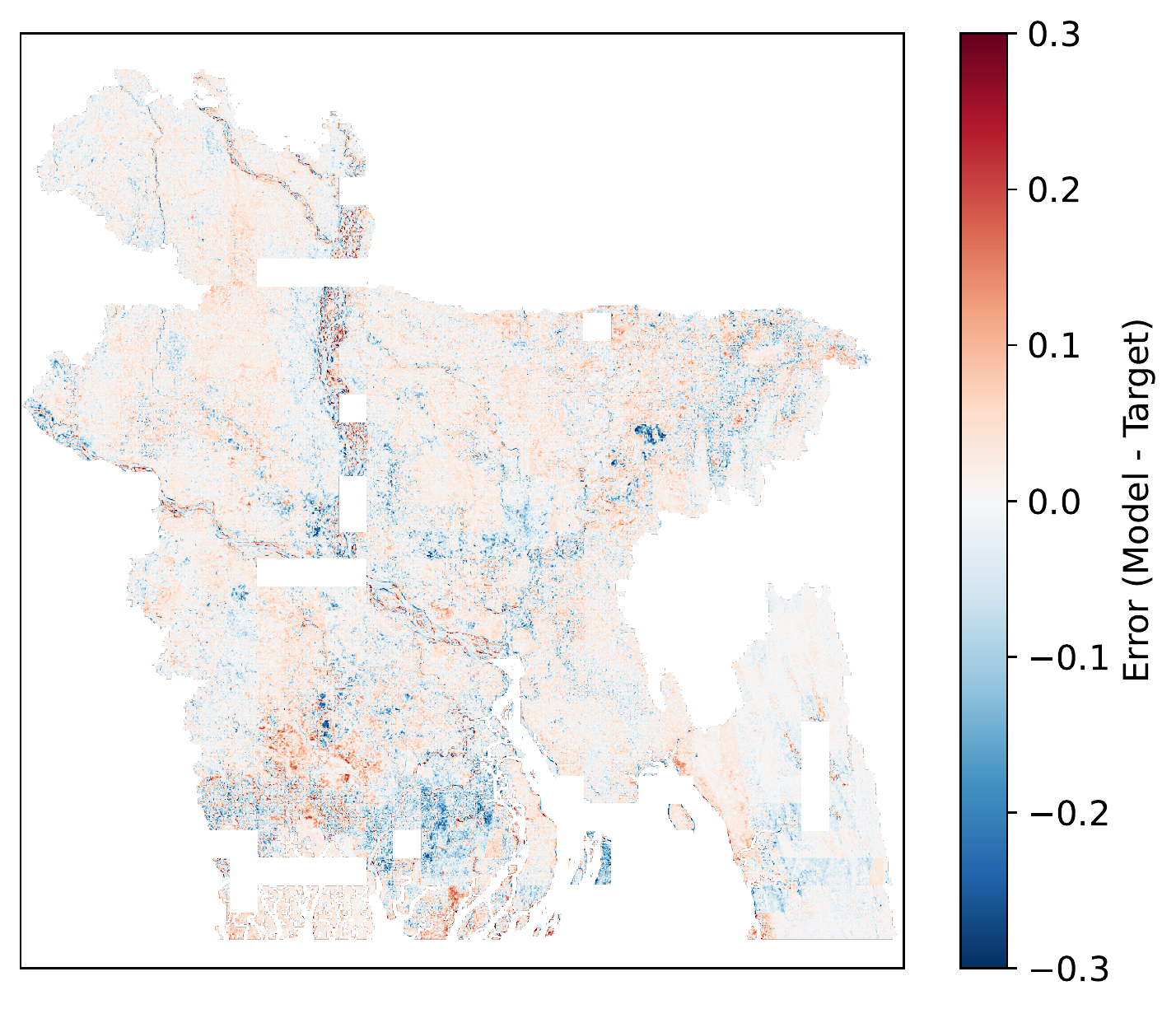}
    \caption{
        Cumulative per pixel error (difference between model and target output) averaged over the cross-validated years for the whole country.
        The blue tint indicates the model under-estimating the inundation area, and the red tint an over-estimation.
        Note that chips that appeared less than 5 times over the 5 years were removed from this figure to avoid visual clutter.
    }
    \label{fig:res:bangladeshPerPixelStat}
\end{figure}
\subsection{CNN-LSTM Model Performance}
    \cref{fig:res:stats} shows performance metrics of the CNN--LSTM aggregated over Bangladesh for the training and testing sets, for all the leave-out years.
    The statistics are consistent amongst cross-validated years.
    The R$^{\text{2}}$ (on average 0.66), slope (on average 0.68) and Spearman's $\rho_s$ (on average 0.69) all depict the model as explaining a large portion of the observed variance, and follow a positive, linear and monotonic relation between the observed and predicted inundation values.
    The model deviates from the mean prediction with a RMSE of 0.1, \ie on average, the model is within 10\% of the observed inundation value.

    The training is displayed along the testing to assess whether the model is overfitting.
    For most years this does not seem to be the case, but 2018 could display signs of overfitting.
    To assess the difference between model fitting and possible overfitting, the ``worst'' and ``best'' year (in terms of testing R$^{\text{2}}$), aggregated over the country, are shown in \cref{fig:res:BangladeshWorstBestYearTimeSeries}.
    In this figure, comparing 2018 when the year is excluded (``worst'') and included (``best'') from the training dataset, the model seems to slightly overfit when the year is excluded.
    In the ``worst'' year, the model appears to overestimate the flood peaks, and underestimate the dry season.
    The same does not occur when 2020 is excluded (the ``best'' year).

    Generally, the model reproduces the peaks during the monsoon season, and valleys during the dry season with a high fidelity.

    \cref{fig:res:bangladeshPerPixelStat} shows the error between the model output and the target (model~-~target) fractional inundated area.
    A majority of the error seems to be located around Bangladesh's major rivers leading to the ocean, where the delineation of water and land is rendered more difficult by the river's meanders.
    The model does not seem to be systematically over or underestimating the inundation extent, but rather simply unable to accurately model the intricate and evolving river extents. Additionally, close to the ocean, in the southern part of the country, as well in the more hilly regions (west side), the model underestimates the fractional inundation area. A majority of the territory shows a negligible average error.

\subsection{Inference for Historical Inundation}
    The goal of the model is to infer the historical inundation extents over Bangladesh.
    \cref{fig:res:ensembleInference} shows the inference over 20 years of MODIS data of the ensemble of cross-validated models.
    \begin{figure}[htb]
    \setcounter{figure}{7}
    \centering
    \includegraphics[width=1\linewidth]{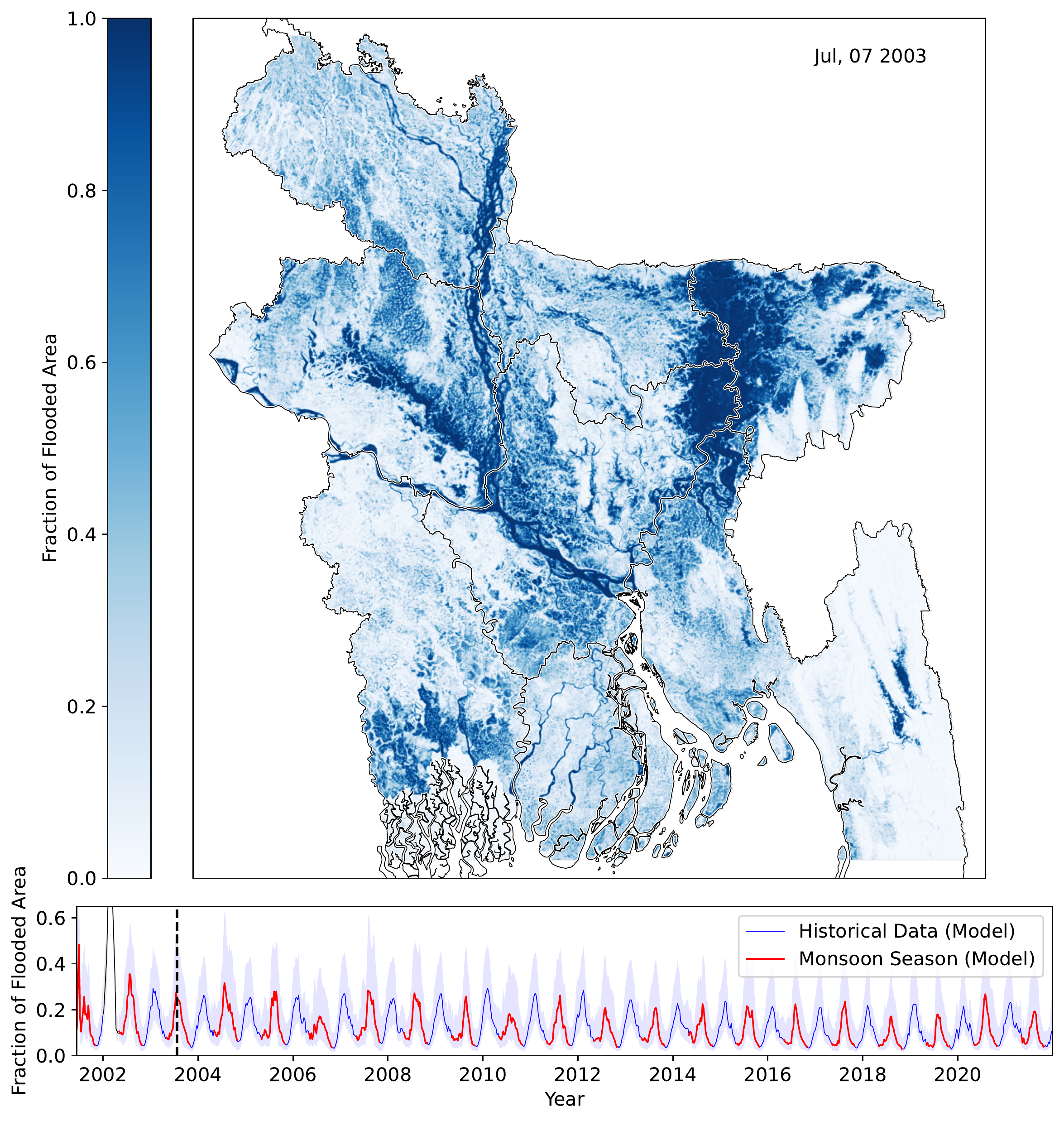}

    \caption{
        Inference run on 20 years of MODIS data (ensemble of the 5 cross-validated models).
        Example of inundation extent for the monsoon flood peak in 2003 over Bangladesh and historical time series, with the monsoon season highlighted in red.
        Note that spring of year 2002 is removed from the time series due to an error in the MODIS data that is propagating through the model with the LSTM.
    }
    \label{fig:res:ensembleInference}
\end{figure}
    The raster image shows an example of inundation at the peak for the year 2003 (July 7$^{\text{th}}$).
    The irrigation inundation peaks occurring during spring are separated from the monsoon inundation peaks.

\begin{figure}[tp]
    \centering
    \includegraphics[width=1\linewidth]{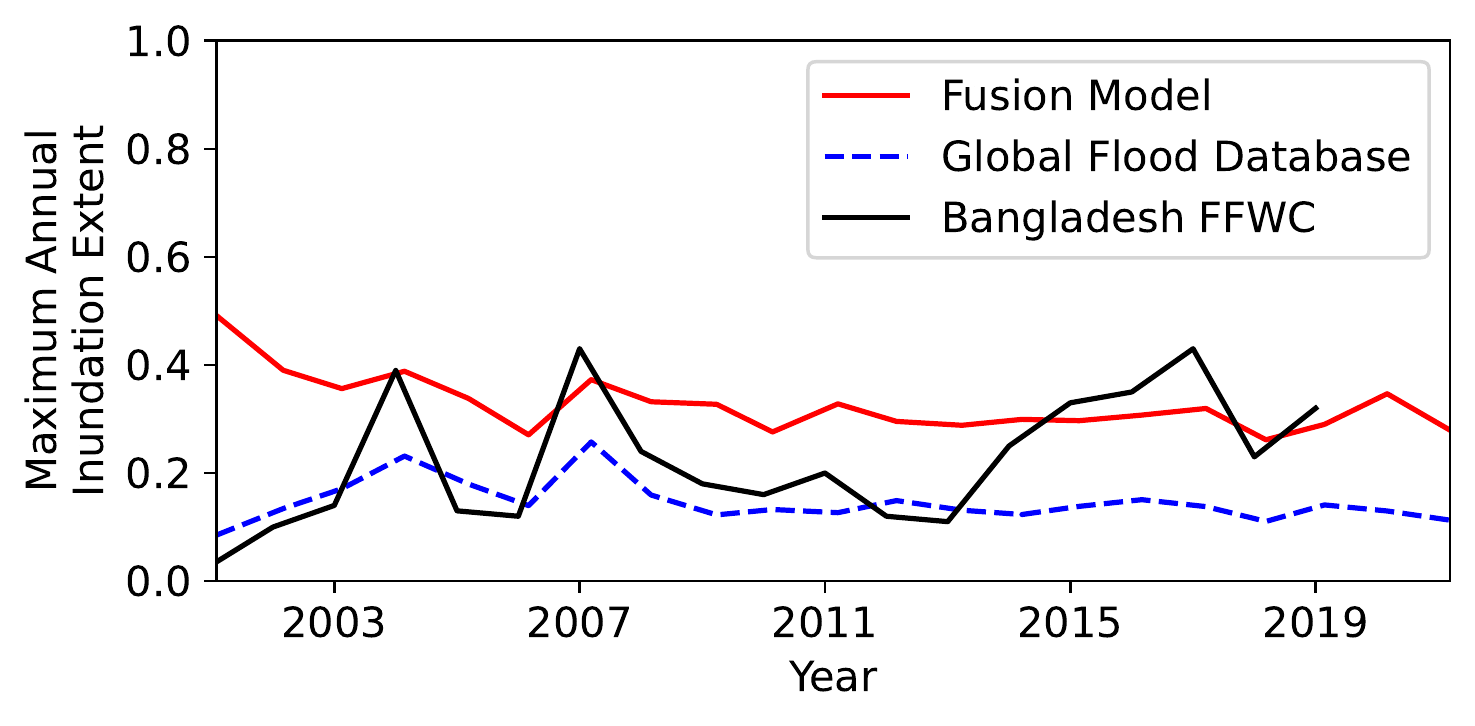}

    \caption{
        Annual monsoon maximal inundation extent extracted from the inferred ensemble CNN--LSTM model (red line, \cf \cref{fig:res:ensembleInference} for the full time series), Global Flood Database (GFD~\cite{tellman2021b}) algorithm and the Bangladesh Flood Forecast Warning Center (FFWC) predictions (based on the MIKE11 Model from the Danish Hydraulic Institute (DHI)).
    }
    \label{fig:res:maximumAnnual}
\end{figure}

    \cref{fig:res:maximumAnnual} shows the annual maximum inundation extent extracted from the fusion model, and the output of two other algorithms for comparison: the global flood database algorithm (GFD~\cite{tellman2021b}) and the Bangladesh Flood Forecast Warning Center (FFWC) prediction (based on the MIKE11 Model from the Danish Hydraulic Institute (DHI)).
    GFD is based on MODIS daily 250 meters resolution images, where each pixel gets a binary inundation prediction.
    The model output is aggregated to 500 meters, the fractional inundated area is thus the average over 4 pixels, then aggregated over all of Bangladesh (mean).
    Mike11's model is a physically based model, the accuracy of which, like all flood models, is affected by large uncertainties, mainly related to the DEM~\cite{bates2022}.
    The DEM in Bangladesh is particularly inaccurate in part because of poor interpolation techniques~\cite{rabby2020}, as well as dynamic seasonal changes to the elevation due to the monsoon filling the water table~\cite{steckler2010} and subsidence due to sediment dynamics and climate change~\cite{steckler2022}.
    The flood map from such a model is not ground truth, but provides a useful reference to compare to remotely sensed flood maps as an independent data source when no ground truth is available, an approach used in other remote sensing flood studies~\cite{cohen2019}.
    Comparisons with MIKE11 and Sentinel-1 flood maps, as well as conversations with the Bangladesh Flood Forecasting and Warning Center that calibrates the model and provided the data, indicate the Mike11 model likely over predicts flood peaks.
    Future work should validate the extracted inundation maps with recent high resolution remotely sensed data, \eg, Planetscope.

    For all models, the inundation peaks are extracted during the monsoon season.
    The monsoon season's inundation peak are uncontrolled (as opposed to controlled during the irrigation season), and of interest for insurance purposes, but for model optimisation it is of interest to use all available inundation observations, \ie, include the irrigation season as well.

    The fusion model replicates all major flood peaks shown in MIKE11 (2004, 2007, 2011 and 2017), whereas the GFD seems to miss some of them.

    In terms of baseline, the fusion algorithm and GFD have, on average, $\sim$0.2 fractional inundated area difference.
    The GFD algorithm is very conservative in terms of water detection, whereas the fusion model here aims at replicating the 10 meters observed fractional flooded area.

\section{Discussion}
\label{sec:discussion}
The temporal features in the deep learning network appear to strongly improve identifying inundation and filling in gaps under clouds, whereas the pure CNN approach seems to struggle in some cases.
The CNN--LSTM likely picks up the ``hydrograph'' signature of the inundation, \ie comprehend the rising and falling of the inundation level and predict whether the water is receding or not in the next time step.
Most importantly, in 2020, which is the highest inundation year in the time series available with Sentinel-1, the LSTM makes the largest contribution, increasing accuracy from 0.63 to 0.72, \ie an additional 10 percent.
Large flood years are especially critical to predict correctly as, for example, in index based insurance applications, large flood years are the years where payouts have to be made by insurers.

Model performance is highest in places with seasonal inundation, where a temporal feature can be learned.
For example in Sylhet, a region in the north east of the country, the seasonality of the lakes filling and recessing are relatively predictable.
Here is where the LSTM shines.
In contrast, coastal areas, areas with steep topography (\ie mountains) and areas prone to storm surges appear to be the least well represented.
This could either be because in these areas, a slow onset of the event, required to generate a temporal feature to fill in the flood gaps, is missing, or because Sentinel-1 (which has a 2 and 10 revisiting period) does not capture storm surge events well along the coast, and fast moving floods in mountainous regions.

Around rivers, where the meanders have changed multiple times over the years, and the river course isn't predictable from one year to the other, the LSTM struggles.
In these cases, a more complicated CNN could learn from more convoluted patterns, and could be coupled with the LSTM framework.

Overall, the CNN--LSTM model proposed here shows an impressive improvement over the CNN-only approach, and proves essential to capture inundation dynamics in time, in regions where the CNN's pattern recognition isn't enough and learning from the past is required.

\section{Conclusion}
\label{sec:conclusion}
The recent failure of Sentinel-1B~\cite{theeuropeanspaceagency2022}, as well as the planned decommission of MODIS in mid-February 2023~\cite{nasa2023}, have made it apparent that products derived from the fusion between different satellite sources are highly important to reduce the dependency of applications to a single product.
Although this work is based on Sentinel-1 and MODIS, future work will bridge the gap between MODIS and VIIRS, MODIS's successor, as 11 years of overlapping data exists between the two.
Additionally, the proposed framework can be operationalised to predict the inundation area when and where Sentinel-1 isn't available, reducing the dependency on Sentinel-1 for, \eg, insurance payouts.

While impressive, the historical MODIS record of 20 years is still relatively short in terms of hydrological return period estimates.
Future work could take advantage of the impressive Advanced Very High Resolution Radiometer (AVHRR) record available since 1979 and combine it with historical Landsat data to create a harmonised dataset and provide longer inundation return period estimates, although capturing the peak inundation extents would remain a challenge with these products.
Additionally, other SAR data, such as, \eg, ENVISAT or PALSAR, could be considered to extend the observed inundation time series.

The proposed model architecture performs reasonably well given the complexity of the task, but could certainly be improved.
The advantage of the proposed approach, treating CNNs and LSTMs sequentially, means that any CNN architecture could replace the one proposed.
Future work should compare the model to a single LSTM to understand the benefits of combining LSTMs with CNNs.
Additionally, future work should explore other architectures, and compare them to alternative solutions to incorporate time (transformers, stack time stamps as features for a CNN).

Overall, the parametric insurance, along with green climate funds, is expanding in countries like Bangladesh.
To support these activities, research like the one proposed here is not only highly relevant, but crucial to allow for realistic and fair deals for all parties involved.

\section*{Code and Data Availability}
\label{sec:avail}
The code and data for this paper can both be found on GitHub: \href{https://github.com/GieziJo/cvpr23-earthvision-CNN-LSTM-Inundation}{https://github.com/GieziJo/cvpr23-earthvision-CNN-LSTM-Inundation}

\section*{Acknowledgements}
\label{sec:ack}
This work is undertaken as part of the NASA New (Early Career) Investigators (NIP) Program (80NSSC21K1044).
Additionally, funding was provided by the Syngenta Foundation for Sustainable Agriculture InsuResilience project and was supported as part of the Columbia World Project, ACToday, Columbia University in the City of New York.

%%%%%%%%% REFERENCES
{\small
\bibliographystyle{ieee_fullname}

% \bibliography{egbib}
\bibliography{main}
}

\end{document}